\documentclass{article}

\usepackage[final,nonatbib]{nips_2016}

\usepackage[utf8]{inputenc} % allow utf-8 input
\usepackage[T1]{fontenc}    % use 8-bit T1 fonts
\usepackage{hyperref}       % hyperlinks
\usepackage{url}            % simple URL typesetting
\usepackage{booktabs}       % professional-quality tables
\usepackage{amsfonts}       % blackboard math symbols
\usepackage{nicefrac}       % compact symbols for 1/2, etc.
\usepackage{microtype}      % microtypography

\usepackage{graphicx} % Required for the inclusion of images
\usepackage{amsmath} % Required for some math elements 
\usepackage{amssymb}

\usepackage{todonotes}
\usepackage{float}

\newcommand{\bm}[1]{{\boldsymbol{#1}}} % Bold math symbol (use for vectors and matrices)
 
\newcommand{\refig}[1]{{Fig. \ref{#1}}} % Ref to figure
\newcommand{\refsect}[1]{{Section \ref{#1}}} % Ref to section

\renewcommand{\vec}{\bm} % for vectors
\newcommand{\matr}{\bm} % for matrices

\title{Collaborative creativity with Monte-Carlo Tree Search and Convolutional Neural Networks}

\author{
	Memo Akten\\
	Department of Computing\\
	Goldsmiths University of London\\
	\texttt{m.akten@gold.ac.uk} \\
	\And
	Mick Grierson\\
	Department of Computing \\
	Goldsmiths University of London \\
	\texttt{m.grierson@gold.ac.uk}	
}

\begin{document}

\maketitle

\begin{abstract}
We investigate a human-machine collaborative drawing environment in which an autonomous agent sketches images while optionally allowing a user to directly influence the agent's trajectory. We combine Monte Carlo Tree Search with image classifiers and test both shallow models (e.g. multinomial logistic regression) and deep Convolutional Neural Networks (e.g. LeNet, Inception v3). We found that using the shallow model, the agent produces a limited variety of images, which are noticably recogonisable by humans. However, using the deeper models, the agent produces a more diverse range of images, and while the agent remains very confident (99.99\%) in having achieved its objective, to humans they mostly resemble unrecognisable `random' noise. We relate this to recent research which also discovered that `deep neural networks are easily fooled' \cite{Nguyen2015} and we discuss possible solutions and future directions for the research.
\end{abstract}

\section{Introduction}
The aim of this research is to create an autonomous, collaborative, `creative' agent that can sketch images in realtime, while optionally allowing a user to intervene and physically influence the agent, affecting how and what it decides to draw. To be more specific, our objective is to allow the agent to take random actions, and sketch any path (i.e. follow any trajectory) that resembles a predetermined, desired object \textit{class}. E.g. everytime we ask the agent to draw a `cat', it should draw a different picture of a cat. Note, the objective for the agent is not to sketch an image that resembles a \textit{specific target picture}, as in that case, no matter how the agent meanders, it would always converge back to the specified target picture. Similarly, we stay away from \textit{any number of} target pictures and instead turn to \textit{Deep Learning} to exploit latent representations that might be useful in aiding the agent. We also allow a user to optionally interact with the agent, and exert forces on it to push and pull it around, causing the agent to diverge onto a new trajectory that might lead to a new picture --- which still resembles the desired class (e.g. it sketches a different `cat' picture). The output images would all be varied, and in effect be collaborations between the user and the agent. The entire system could be seen as a realtime collaborative drawing environment between a human and a `creative' autonomous agent. This research is presented as work in progress. At this stage we have mixed results (discussed in \refsect{sect:results}) where the system shows potential, but also demonstrates weaknesses relating to the Deep Learning models.

\section{Background}
Computer generated, procedural visual art is a rich field with its roots dating back to the 1950s with John Whitney Sr., followed by artists such as Paul Brown, Vera Molnar, Manfred Mohr, Frieder Nake, Larry Cuba and more from the 1960s onwards \cite{Grierson2005}. Harold Cohen's AARON \cite{Cohen1973}, was arguably the first software using Artificial Intelligence (AI) to produce visual art, followed by artists such as William Latham \cite{Todd1992}, Karl Sims, \cite{Sims1994}, and Scott Draves \cite{Draves2005}. Also in the 1960s and 70s, Myron Kruger introduced interactivity --- particularly \textit{realtime gestural interactivity} --- into computer artworks, culminating in his seminal Artificial Reality \textit{Responsive Environment} `Videoplace' \cite{Krueger1985}. More recently, works such as Simon Colton's `Painting Fool' \cite{Colton2012} and Patrick Tresset's `Artistically Skilled Embodied Agents' \cite{Tresset2014} also investigate the 'psychology' of `creative' AI agents.

While this is the context for our research, computational models of creativity --- such as those described in \cite{Boden1998} --- or methods of evaluating machine creativity --- such as those described in \cite{pease2001evaluating} --- are currently not within the scope of this study. Instead, we focus on investigating the opportunities provided by Deep artificial Neural Networks (DNNs) --- in particular \textit{Convolutional Neural Networks (CNN)} --- combined with agent-based AI methods --- such as \textit{Monte Carlo Tree Search (MCTS)} --- for the task of driving collaborative drawing agents. 

Deep convolutional architectures saw initial success in image classification in the late 1980s \cite{LeCun1989}, and they are now consistently providing state of the art results in this field \cite{Krizhevsky2012,Szegedy2014,He2015,Szegedy2016}. An active area of research is focused on trying to understand and visualise the internal workings of CNNs, and use them for generative purposes \cite{Zeiler2013,Mahendran2014,Mahendran2015,Radford2015,Mordvintsev2015,Gatys2015}. Additionally, CNNs have also been used with great success to drive AI agents, such as those playing Atari games, Doom or Go \cite{Mnih2013,Guo2014,kempka2016vizdoom,Silver2016}. In these latter cases, the CNNs have effectively acted as the agent's `eyes', breaking down the raw input (i.e. screen pixels or state of the Go board) into more abstract, meaningful representations which can be used by the agent. These researchers have also used MCTS \cite{Browne2012} to guide the agents, and allow them to make more optimical decisions based on the meaningful representations provided by the CNN. It is based on this direction of research, and within the artistic, `artificially creative' context outlined above that our study takes place.

\section{Method}
The agent, similar to a LOGO Turtle \cite{Papert1980}, can move forwards (leaving a trail as it moves), or rotate clockwise or counter-clockwise. We use MCTS to simulate many alternative actions and paths at every timestep, and at the end of each simulation, the entire trajectory of the agent --- including the previous history as well as the simulation --- is fed into an image classifier. The classifier evaluates how much the agent’s trajectory resembles the desired class, and that score is used as the reward backpropagated by MCTS to choose the final action. This is described in more detail in the sections below.

We tested three classifiers trained on two datasets: i) the MNIST dataset of hand-written digits \cite{LeCun2010}, and ii) the ImageNet dataset of millions of labeled images classified into one thousand classes\cite{Deng2009}.

We developed our application in \textit{openFrameworks} \cite{openframeworks}, a C++ creative development toolkit very popular amongst artists and creative coders. We used \textit{ofxMSAmcts}, our own open-source C++ MCTS implementation (i.e. \textit{addon}) for openFrameworks  \cite{Akten2015-ofxMSAmcts}. We built, trained and saved the image classifiers in python using TensorFlow \cite{tensorflow2015}, and loaded the trained models into  C++/openFrameworks using \textit{ofxMSATensorFlow}, again our own open-source TensorFlow openFrameworks wrapper (addon) \cite{Akten2016-ofxMSATensorFlow}.

\subsection{MCTS Agent}
The MCTS procedure is summarised in Appendix \ref{mcts_theory} and a more detailed survey can be found in \cite{Browne2012}. In our implementation, our agent can move forward with a number of different speeds ($n_{speeds}$) to choose from, ranging from $s_{min}$ to $s_{max}$. It can rotate in $r_{inc}$ increments ranging from $-r_{max}$ to $+r_{max}$, giving it $n_{rot} = {2r_{max}}/{r_{inc}}$ number of different rotations to choose from. Thus at each timestep, the agent has $n_a = n_{speeds} * n_{rot}$ total number of actions to choose from. An example configuration is: $n_{speeds}:=2, s_{min}:=0, s_{max}:=2, n_{rot}:=7, r_{max}:=30, \quad \text{with} \quad n_a = 14$.

The state $s$ at any timestep, is the full trajectory (stored as a vector path) that the agent has followed (i.e. sketched) up to that point in time. The state $s'$ which is reached after the agent takes an action $a$, is the full image that will be drawn once the agent takes that action. The MCTS simulations end after a maximum rollout depth (e.g. $D = 100$), after which the final simulated drawn image (rasterized as a bitmap image) is fed into an image classifier, and the probability score for the desired class is retrieved and backpropagated as the reward for reaching that state. In our application we also visualise all of the `imagined' paths (i.e. all simulations). 

\subsection{MNIST - Multinomial logistic regression}
The first classifier we train is a \textit{multinomial logistic regression} trained on the MNIST dataset, a linear transformation followed by a softmax activation \cite{Bishop2006}. We formulate this as

\begin{equation}
\vec{y} = softmax(\matr{W}\vec{x} + \vec{b}), \quad \text{where} \quad softmax(x)_i = \frac{e^{x_i}}{\sum_j{e^{x_j}}}, \\
\end{equation}

$\vec{x}$ is a vector containing the flattened input image pixels, and $\matr{W}$ and $\vec{b}$ are respectively the \textit{weights matrix} and \textit{bias vector} to be learnt through training. We train using stochastic gradient descent and backpropagation trying to minimise the cross-entropy \cite{Bishop2006} between $\vec{y}$, the predicted probability distribution of our model, and $\vec{y'}$, the true distribution, i.e. the training data.

\begin{equation}
H(\vec{y}', \vec{y}) = - \sum_i y'_i log(y_i)
\end{equation}

The model completed training in a few seconds (on CPU) and unsurprisingly it doesn't generalise very well, scoring about 90\% on the validation data. As expected, it is very restrictive on the types of inputs it can classify and does not provide much translational, rotational or scale invariance as the softmax operates directly on a linear transformation of the flattened raw pixel data.

\subsection{MNIST - Convolutional Neural Network (LeNet)}
The second classifier we train is a deep Convolutional Neural Network (CNN) similar to the LeNet architecture \cite{LeCun1998}. CNNs are deep architectures inspired by the visual cortex, and consist of a hierarchy of \textit{stacked feature maps}, where each layer extracts features from the layer below using two-dimensional convolution kernels which are also learnt during training. We use two convolutional stacks, each stack consisting of 5x5 convolution kernels followed by an element-wise rectified linear unit ($ReLU(x) = max(0, x)$ \cite{Jarrett2009}), followed by a 2x2 max-pool layer. The first convolutional stack learns 32 feature maps while the second stack learns 64 feature maps. After the two convolution stacks, we use a dense fully connected layer with 1024 neurons, followed by a softmax to convert the results to a normalised probablity distribution. Similar to the multinomial logistic regression, we train with stochastic gradient descent and backpropagation to minimise the cross-entropy. We use dropout regularization with a probability of 50\% \cite{HintonDropout2014}. 

The model trained in a couple of minutes (again on CPU). It scored 99.2\% accuracy on the validation data and unsurprisingly proved to be much more successful in classifying images, with higher resilience to noise, translation, scale and rotation compared to the multinomial logistic regression.

\subsection{ImageNet - Inception-v3}

For the final classifier, we download and use a pre-trained model of Google's state of the art image classification architecture Inception-v3 \cite{Szegedy2015}, trained on ImageNet. This is a deep CNN reaching as low as 3.58\% error in the 2012 ImageNet validation set for Top-5 error.

\section{Results and conclusion}
\label{sect:results}
The overall system architecture seems to work and shows potential.

\begin{figure}[H]
	\includegraphics[width=\textwidth]{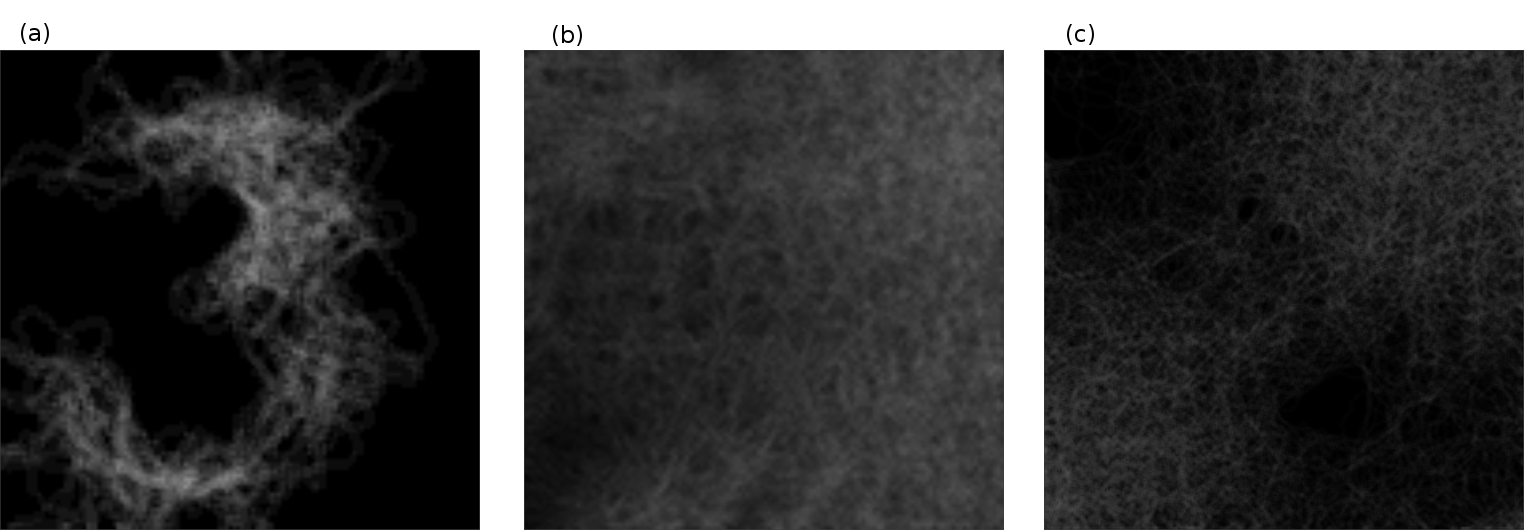}
	\caption{Agent trying to sketch (a) the digit `3' using Multinomial Logistic Regression trained on MNIST as the classifier, (b) a `meerkat' using Google's pre-trained state of the art Inception-v3 model as image classifier, (c) a `wolf' using the same system. Interestingly, the output of the latter two looks like random noise to us, but the classifier evaluates both images with extremely high confidence to be the desired class. See \refig{fig:mcts_3} for full screenshot including confidence values and simulations.}
	\label{fig:mcts_all}
\end{figure}

Using MNIST with the shallow model --- multinomial logistic regression --- proved very effective and positive results were obtained. Example output can be seen in \refig{fig:mcts_all} (a), where the agent has clearly sketched the desired image which was the digit `3'. Running this multiple times with different objectives provided similar results, demonstrating that the MCTS + Image Classifier system works on the whole.

However, we found that the deep CNNs were not ideal for generative applications in this manner, as they provide too many false positives. As discriminative models trained to classify images, they can classify \textit{natural} images \textit{correctly} with very high accuracy and confidence, but they also \textit{incorrectly} classify \textit{unnatural} images such as noise and abstract shapes, with equally high confidence. In other words, the class manifolds successfully represent the desired \textit{natural images}, but also include \textit{undesired unnatural images} as well. In retrospect this is not too surprising, as these networks are designed to function under very high noise with high degree of translational, rotational and scale invariance, and the training data does not contain any unnatural images.

Ultimately, the tendency of these deep models to identify false positives, cause the agent to wander around the screen randomly, sketching out images that appear to humans as unrecognisable `random' noise, but are classified by the network with very high (99.99\%) confidence to be the desired class. This can be seen in \refig{fig:mcts_all} (b) and (c). Interestingly, even though the system produces what appears to humans as random noise, it's not only the \textit{top} prediction of the network which classifies the outputs as the desired class, but the \textit{top 5} predictions as well. E.g. in \refig{fig:mcts_all} (b) the desired class is `meerkat'. The top 5 predictions of the network are `meerkat' (99.999\%), `mongoose', `hyena', `Egyptian cat', `cheetah' --- all small, cat-like animals. In \refig{fig:mcts_all} (c) the desired class is `white wolf'. The top 5 predictions are `white wolf' (97.8\%), `timber wolf', `Arctic fox', `Samoyed', `west highlands white terrier' --- all white/light grey dogs.

In summary, the failure of the system --- i.e. the output not resembling the desired class in our eyes --- is not a failure of the \textit{whole} system, or of MCTS, or the way MCTS and the classifier are integrated. But it is a failure of the deep CNN classifier, in that it's easily susceptible to false positives. This also confirms recent research that deep CNNs can easily be fooled \cite{Nguyen2015}. Though this previous research demonstrated this using evolutionary algorithms, and in a very non-realtime, non-interactive manner, as opposed to our realtime agent based methods. There are also recent examples of overcoming this problem of generating \textit{unnatural} images, by using \textit{natural image priors} \cite{Mordvintsev2015,Nguyen2016} or adversarial networks \cite{Goodfellow2014a,Radford2015}. Both approaches we are planning to try in future research. 

At every timestep, we also visualise all of the MCTS rollouts (\refig{fig:mcts_3}). From a conceptual and aesthetic point of view this can be thought of as visualising the possible trajectories `imagined' by the agent, which is in part the motivation for this research.

\small
\bibliography{msa.bib}

\clearpage
\appendix

\section{Monte Carlo Tree Search}
\label{mcts_theory}

\begin{figure}[H]
	\includegraphics[width=\textwidth]{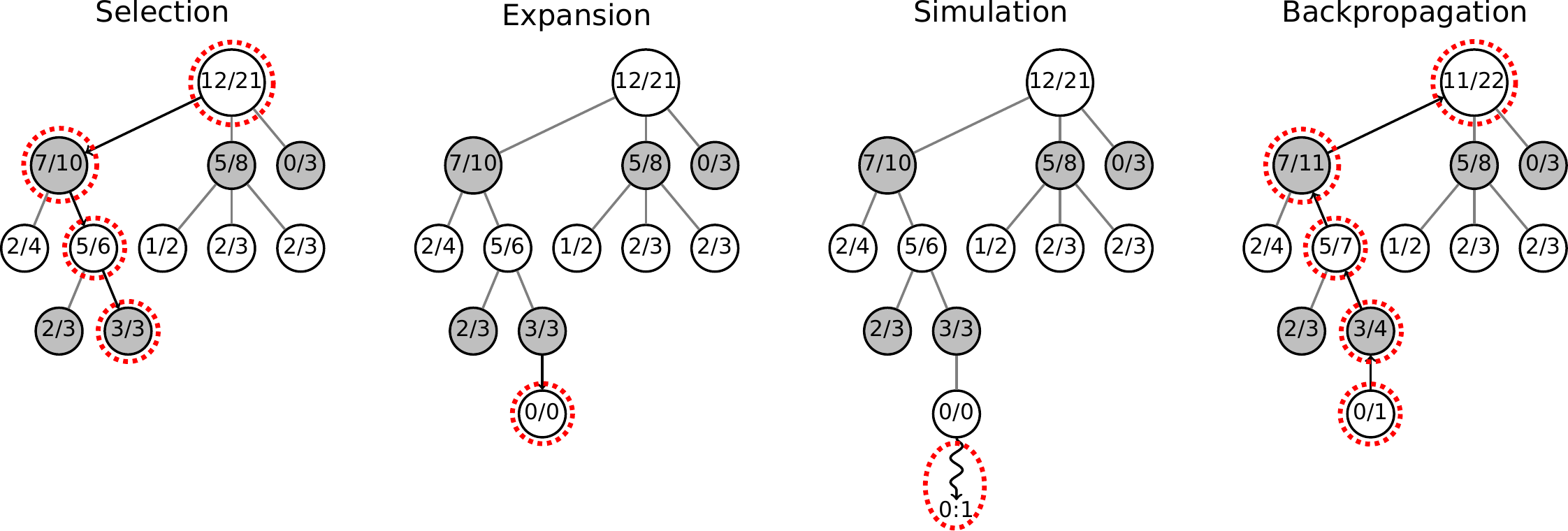}
	\caption{Overview of MCTS. Image from wikipedia, CC BY-SA 3.0 }
	\label{fig:mcts_overview}
\end{figure}

MCTS is a probabilistic approach to searching a decision tree. At its core, it treats actions like slot machines in a multi-armed bandit. Below we briefly discuss our implementation. A detailed explanation and survey can be found in \cite{Browne2012}.

We represent our problem as a discrete-time \textit{Markov Decision Process} (MDP) \cite{bellman1957markovian}. At a particular timestep the system is in a \textit{state} $s$. The agent can take an \textit{action} $a$ (out of a number of available actions $n_a$) to arrive at a new state $s'$, and will be given a \textit{reward} $R_a(s, s')$ for doing so. Instead of trying to simulate every possible action at every timestep, MCTS uses a heuristic to only partially expand what might be the most promising branches of the decision tree. At every timestep, many simulations are run, but only from carefully chosen nodes of the decision tree. The algorithm decides which node to expand and runs a simulation from it based on a balance between \textit{exploiting} known rewards for nodes already fully expanded, vs \textit{exploring} nodes which have not yet been fully explored, i.e. with unknown rewards. 

An overview of MCTS can be seen in \refig{fig:mcts_overview} and summarised as i) \textit{Select} a node to expand by starting at the current root node and recursively stepping through fully expanded child nodes until an \textit{expandable node} is reached (i.e. a non-terminal node with unvisited children), ii) \textit{Expand} the selected node (i.e. pick one of its unvisited children), iii) Run a \textit{simulation} (i.e. \textit{rollout}) from the new child node until an \textit{outcome} is reached, iv) \textit{Backpropagate} a reward value $\Delta$ based on the outcome of the simulation, back up the tree, accumulating statistics for each node.

At each timestep, this procedure of \{\textit{selection-expansion-simulation-backpropagation}\} is repeated many times to partially expand the decision tree, until a predefined criterion is met. This is usually an allocated time budget (e.g. maximum milliseconds per timestep) or an upper bound on the number of simulations per timestep, or a domain-specific interruption criterion. Once the search is terminated, a final decision is made by choosing an action according to a criterion based on the search statistics. E.g. The most visited child node (\textit{Robust Child}) or child with highest rewards (\textit{Max Child}), more examples can be found in \cite{Browne2012}.

The \textit{tree policy} decides which node to select and expand, and we use the \textit{Upper Confidence Bound 1 applied to Trees} (UCT), formulated as

\begin{equation} \label{eq:uct}
\frac{v}{n_c} + k \sqrt{\frac{2\ln{(n_t)}}{n_c}}
\end{equation}

where $v$ is the \textit{value} of the child node, $n_c$ is the number of times the child node has been visited, $k$ is the \textit{exploration parameter} (theoretically equal to $\sqrt{2}$ but usually chosen empirically) and $n_t$ is the total number of simulations for the node considered. The \textit{default policy} decides how to choose actions during the simulation (i.e. rollout). We use a \textit{random rollout}, i.e. actions are selected randomly from the $n_a$ available actions.

In a typical two-player zero-sum game such as tic-tac-toe, MCTS simulations will run until they reach a \textit{terminal state}, where the outcome  is either a \textit{win}, \textit{lose}, or \textit{draw}. In these situations, usually a simple reward value $\Delta = 1$ is backpropagated if the agent wins, $\Delta = -1$ if the agent loses, and $\Delta = 0$ if the result is a draw. Since each simulation iteration combinatorially increases the size of the decision tree, in situations where the tree is very \textit{deep} --- such as in the game of Go \cite{Silver2016} --- it is not feasible to run simulations until a terminal state is reached. In these cases it is common to prematurely end the simulation after a predefined \textit{maximum rollout depth $D$}, and then evaluate or approximate the reward value $\Delta$ using a heuristic on the simulation end-state. This approximated reward value is then backpropagated as before.

\section{Screenshots}
\begin{figure}[H]
	\includegraphics[width=\textwidth]{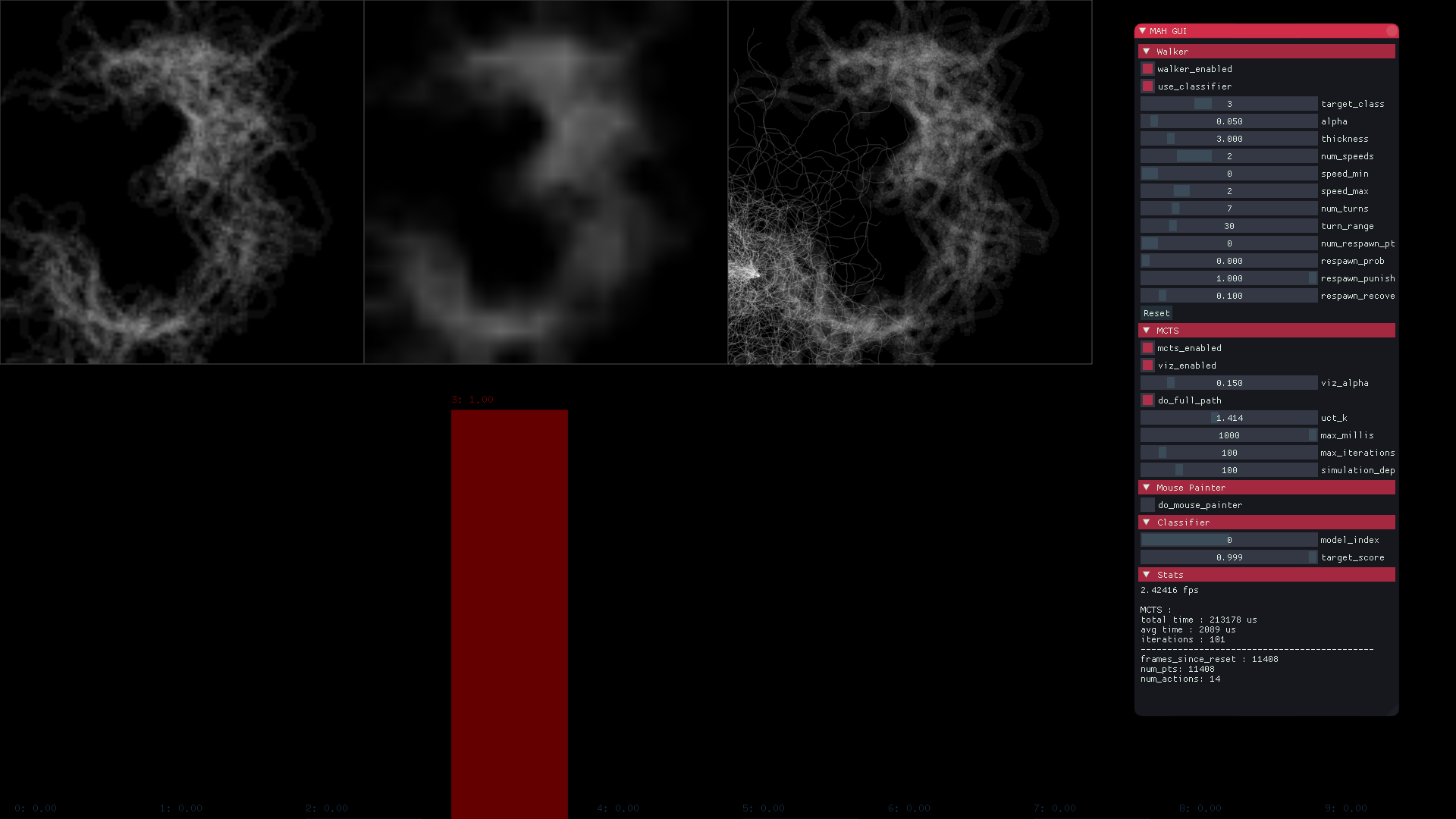}
\end{figure}

\begin{figure}[H]
	\includegraphics[width=\textwidth]{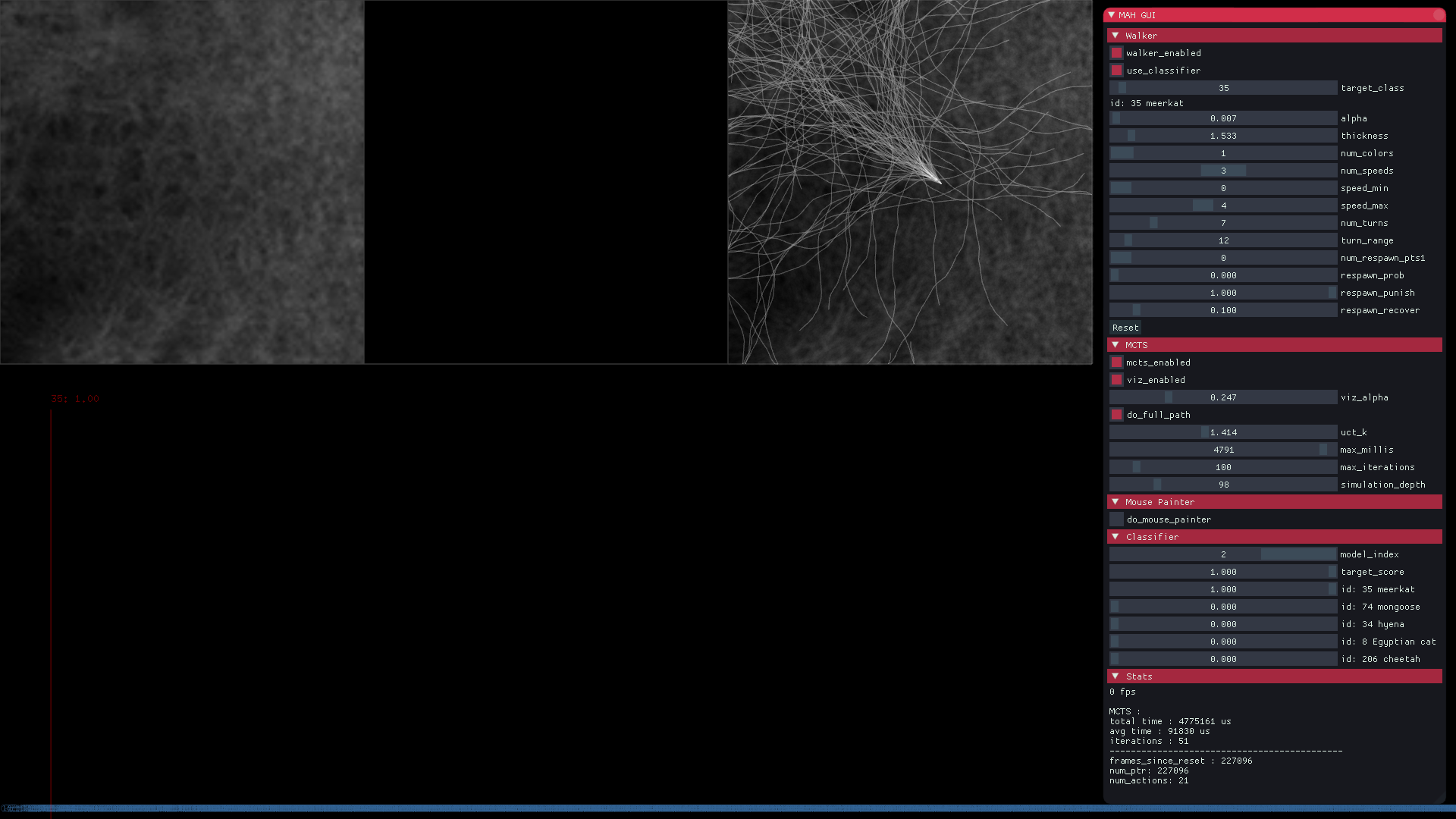}
	\caption{
		Full screenshots for agent sketching and simulation software. Top: sketching the digit '3' using the Multinomial Logistic Regression trained on MNIST as the classifier, bottom: sketching a `meerkat' using Google's pre-trained state of the art Inception-v3 model as image classifier. In both instances the top-left viewport shows the current output, top-right viewport shows a visualisation of the current MCTS rollouts (showing all simulated paths which will be evaluated by the classifier.}
	\label{fig:mcts_3}
\end{figure}

\end{document}